\documentclass[a4paper]{article}

\usepackage{INTERSPEECH2019}
\usepackage{multirow}
\usepackage{soul}
\usepackage{subcaption}
\usepackage{hyperref}

\title{Fast and Accurate Capitalization and Punctuation for \\ Automatic Speech Recognition Using Transformer and Chunk Merging}
\name{Binh Nguyen$^{1,5}$, Vu Bao Hung Nguyen$^1$, Hien Nguyen$^{1,3}$, Pham Ngoc Phuong$^{1,3}$, The-Loc Nguyen$^{1,4}$, Quoc Truong Do$^1$, Luong Chi Mai$^{1,2}$}
\address{
  $^1$Vietnam Artificial Intelligence System, Vietnam\\
  $^2$University of Science and Technology of Hanoi, Vietnam\\
  $^3$Thai Nguyen University, Vietnam\\
  $^4$Hanoi University of Mining and Geology, Vietnam\\
  $^5$Hanoi University of Science and Technology, Vietnam
}
\email{\{binhnguyen|hungnguyen|locnguyen|truongdo\}@vais.vn, nguyenthuhien@dhsptn.edu.vn, phuongpn@tnu.edu.vn, lcmai@ioit.ac.vn}

\begin{document}

\maketitle
\begin{abstract}
In recent years, studies on automatic speech recognition (ASR) have shown outstanding results that reach human parity on short speech segments. However, there are still difficulties in standardizing the output of ASR such as capitalization and punctuation restoration for long-speech transcription. The problems obstruct readers to understand the ASR output semantically and also cause difficulties for natural language processing models such as NER, POS and semantic parsing. In this paper, we propose a method to restore the punctuation and capitalization for long-speech ASR transcription. The method is based on Transformer models and chunk merging that allows us to (1), build a single model that performs punctuation and capitalization in one go, and (2), perform decoding in parallel while improving the prediction accuracy. Experiments on British National Corpus showed that the proposed approach outperforms existing methods in both accuracy and decoding speed.
\end{abstract}

\section{Introduction}
In a typical setup of an ASR system, punctuation and capitalization of words are
removed because they do not affect the pronunciation of words.
As the result, the output of ASR contains purely a sequence of words or
alphabet characters depending on the model type. While this output is
sufficient for many applications, such as voice commands, virtual assistants,
where speech segments are usually short and independent,
it is difficult to be used in applications that transcribes long speech
segments. It would be easier for human to read a document with
proper punctuation and word capitalization. 
Moreover, when ASR results are fed into NLP models to perform
machine translation (MT) or name entity recognition (NER), punctuation and
word capitalization are crucial pieces of information that can help to boost
the performance \cite{cho2012segmentation,tkachenko2012named,yadav2018survey}.

\begin{figure}[t]
  \centering
  \includegraphics[width=\linewidth]{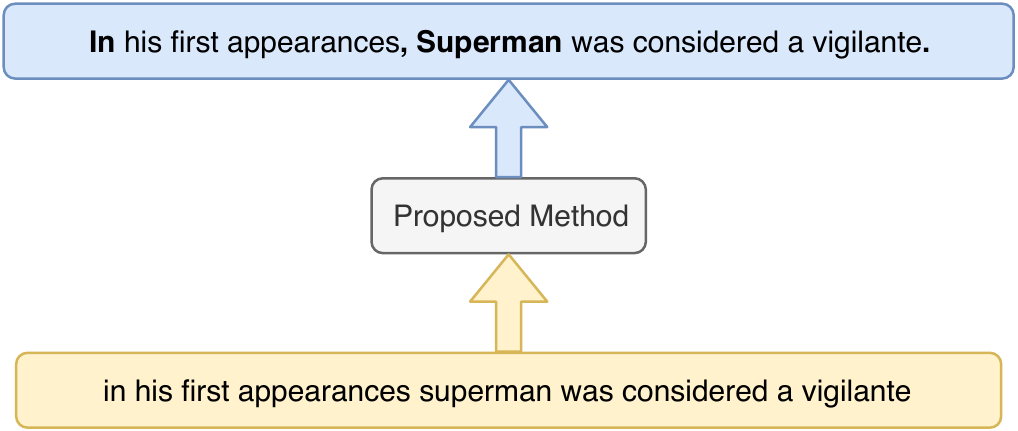}
  \caption{The proposed method for performing both punctuation and word capitalization in one go}
  \label{fig:overview}
\end{figure}

Regarding studies on segmentation and punctuation insertion for ASR, 
Cho et al. \cite{cho2012segmentation} proposed a method to use phrase-based translation
models that consider the punctuation insertion as machine translation tasks. 
The model takes input is unpuctuted text and translates into a punctuated one.
Zelasko et al. \cite{_elasko_2018} and Tilk et al. \cite{tilk2015lstm} 
incoporate more information from speech signal to improve the performance.
In \cite{lu2010better, 
ueffing2013improved}, dynamic conditional random fields (CRFs) 
\cite{lafferty2001conditional} were used to predict punctuation.
The works proposed by Cho et al. \cite{cho2017nmt} and Tilk et al. \cite{tilk2015lstm} 
made use of end-to-end translation model with LSTM to predict
punctuation and segmentation. They successfully demonstrated that the end-to-end models
outperform conventional approaches.
While existing works are capable of predicting
punctuation, they share similar limitation.
First, the models only handle one task which is punctuation insertion, however,
output from ASR is also typically uncapitalized. While adding just punctuation might help
speech translation to determine when to translate, other NLP tasks such as NER and PoS tagging
do not get much help because one of the key feature of these models is word capitalization. 
Second, long input sentences are usually split into fix-length and non-overlapped chunks before feeding
into the model. Although this method helps to speedup the inference by processing chunks independently and in parallel, it is prone to bad prediction of words around the chunk's boundary because there isn't enough both left and right context information in the area.

In this paper, we proposed a method based on transformer models
and overlapped chunk-merging to restore both word capitalization
and punctuation in one go as illustrated in Figure~\ref{fig:overview}. 
The system consists of 3 components (Figure~\ref{fig:system} - b).
The first component is an overlapped chunk spliting that takes a long input sequence
and splits them into chunks with overlap. This process make sure that the second component,
which is the capitalization and punctuation model, always have enough left and right context
of words to make the prediction. The last component is the chunk-merging where the overlapped
output are combined into a single sentence. This process decides which part of the overlap area
to be removed and to be kept.
The method allows us to (1), build a single model that performs punctuation and capitalization
without the need of pipeline results from one system to another,
and (2), perform decoding in parallel while improving the prediction accuracy.

\section{End-to-end Model for Punctuation and Segmentation}



\begin{figure}[t]
  \begin{subfigure}{\linewidth}
      \centering
      \includegraphics[width=\linewidth]{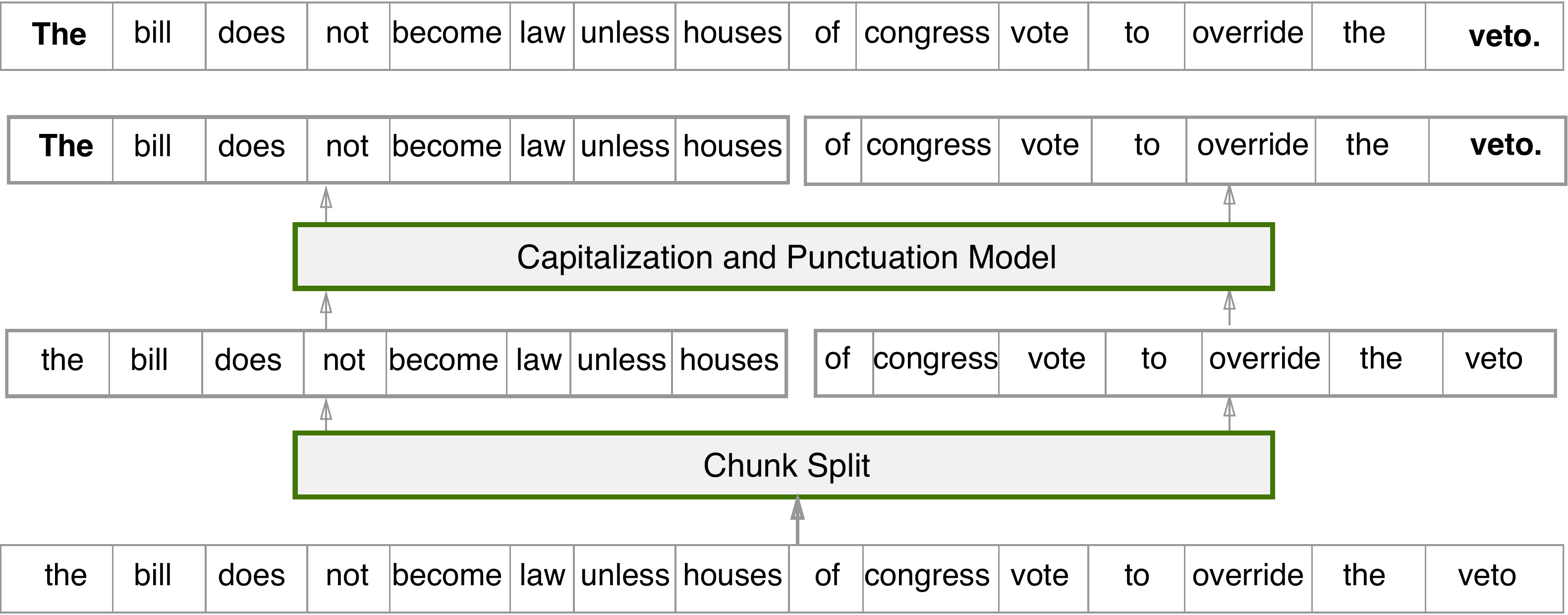}
      \caption{Capitalization and Punctuation System Without Overlapping Segments}
      \label{fig:non-overlap}
  \end{subfigure}
  \begin{subfigure}{\linewidth}
      \centering
      \includegraphics[width=\linewidth]{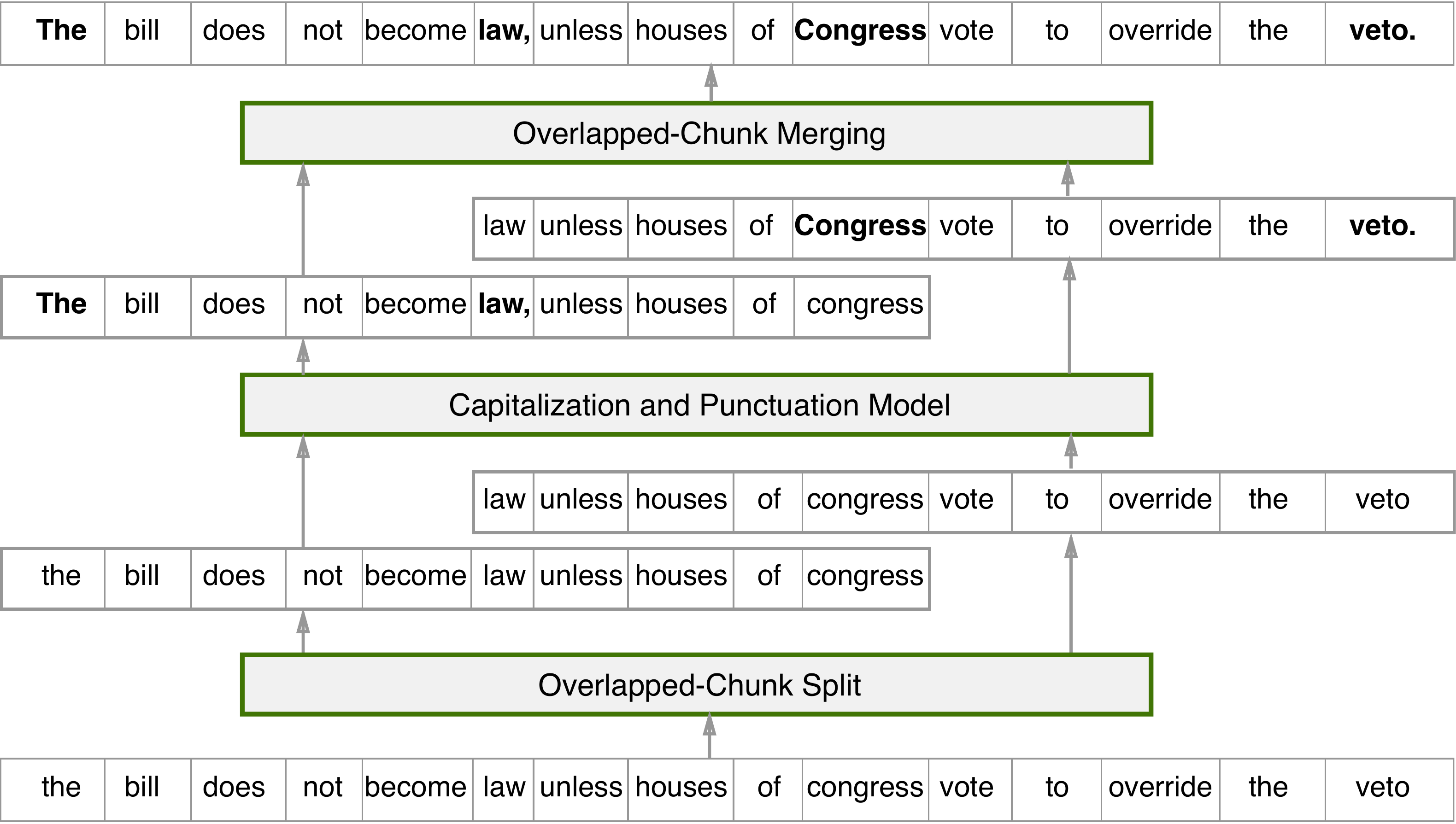}
      \caption{Proposed System Architecture for Capitalization and Punctuation. Because of more context, it can add comma after ``law'' and upper case ``congress''}
      \label{fig:overlap}
  \end{subfigure}
  \caption{Capitalization  and  Punctuation  System With and  Without  Overlap-ping Segments. Ground truth of this example is ``The bill does not become law, unless Congress vote to override the veto.''}
  \label{fig:system}
\end{figure}

End2end models for punctuation works in a similar way with machine translation tasks \cite{opennmt,bahdanau2014neural} where it takes
input is a sequence of of lowercase, unpunctuated words and outputs a sequence with truecase and punctuation inserted.
Figure~\ref{fig:non-overlap}
illustrates the use of end-to-end models for restoring capitalization and punctuation proposed in \cite{sutskever2014sequence}. First, a long input text from ASR is split into small segments 
and then, they are fed into a translation model to produce an output sequence.
While the approach can take advantages of LSTM models that it is able to learn longer context information,
it usually failed to predict truecase or punctuation of words near the segment boundary.

Previous studies \cite{vaswani2017attention} has pointed out that Transformer performs better than LSTM models by exploiting its self-attention layer to capture context more efficiently and speedup the training process.
Transformer is basically an encoder-decoder model. It contains multiple identical encoders and identical decoders stacked upon each other. Each encoder has a self-attention layer that extract surrounding words information when a word is being encoded. This layer is followed by a feed forward neural network; the networks in different encoders do not share weights. Each decoder also has a self-attention layer and a feed forward neural network, but to enhance the relevant parts of input, an attention layer (similar to attention in sequence-to-sequence model) is added between the 2 sub-components.

Transformer's architecture was hand-crafted manually, Evolved Transformer (ET) was created to enhance Transformer. The idea behind ET is using neural architecture search (NAS) \cite{zoph2016neural} to look for the most promising setup among different alternatives of neural networks. To modify Transformer model configuration toward a better one, ET uses an evolution-based algorithm with an innovative approach to expedite the process. 



\section{Proposed Method}
Figure~\ref{fig:overlap} describes our system architecture. The system works as follows,
first, output from and ASR module (lowercase without punctuation) is fed to the Overlapped-Chunk Split module 
 to produce overlapped segments. Second, the Capitalization and Punctuation Model takes
the split segments and processes them in parallel to output a list of outputs. Finally, the outputs are
merged back to form a final sentence using the Overlapped-Chunk Merging module.
Details of each modules are described in the following sections.

\subsection{Capitalization and Punctuation Model}

This section describes the architecture and hyperparameters of our models. To be certain that our method of overlapping segments are efficient regardless of models, we preformed the experiments on sequence-to-sequence LSTM model and Evolved Transformer framework one by one. Our models are implemented based on Tensor2Tensor\cite{tensor2tensor} and OpenNMT\cite{opennmt} framework. Concatenating overlapped chunks is developed as a separated module and used only after the inferring process.

To replicate the same condition, both the models have 6 hidden layers, word embedding size of 256, batch size of 4096 and trained for 200 epochs; the number of head in transformer model is 8. Their jobs is to convert from a sequence of lowercase text without punctuation to another sequence of capitalized text with punctuation.
With 500 MB of text data for training, each model took 20 hours to train on an NVIDIA 2080Ti GPU.




\begin{figure*}[t]
  \centering
  \includegraphics[width=0.65\linewidth]{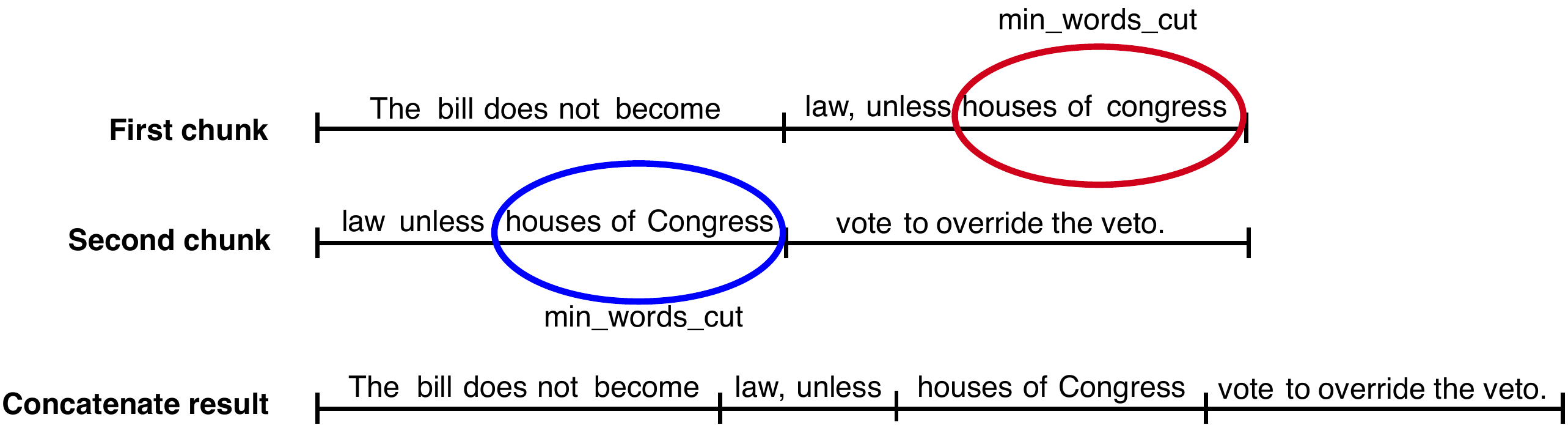}
  \caption{Overlapped Chunk Concatenation}
  \label{fig:min-words-cut}
\end{figure*}

\subsection{Algorithm for Overlapped-Chunk Split and Merging}
From preliminary experiments, we observed that the model often makes
mistakes when processing words near the chunk boundary. We hypothesize that there is not enough context information around the area,
leading to the poor performance of the model. To mitigate the problem, we proposed a method to split long input sentences into chunks with a
chunk size of $k$ words and a sliding window of $k/2$ words so that 2 consecutive chunks are overlapped.
Later, the output of the model are merged in the way that we only keep predictions of the model where there is enough context information (an example is illustrated in Figure~\ref{fig:overlap}).

While splitting input sentences into overlapped chunks is straight-forward as we only need to decide the chunk and overlapped size, merging the overlapped results is more difficult. Since the output of the overlapped
region between 2 consecutive chunks can be different, we need to decide which words to keep and which word to remove to form
a complete sentence.
According to the hypothesis above, we defined a parameter called \verb|min_words_cut| that indicates the number of words at the end the first chunk to be removed and also the number of words to be kept at
the end of overlapped words in the second chunk. It ranges from 0 to the overlap size.
With the value of 0, the whole overlapped words in the first chunk are
kept while the overlapped words in the second chunk are removed (illustrated in  Figure~\ref{fig:min-words-cut}). The 
same principle is applied when \verb|min_words_cut| equals to the
overlapped size.


\subsection{Data Preparation}
To simulate the ASR output, we preprocess the dataset as followed. First, the characters are cleaned up: only the alphabet characters and three punctuation (comma, full stop and question mark) are kept. Then, we make sure that the punctuation belongs to the previous word, for instance, we use ``laptop, mobile'' not ``laptop , mobile''.
Finally, we split data into chunks according to the split algorithm described in the above section. An example is shown in Figure~\ref{tab:data-sample}.

\begin{figure}[t]
    \centering
    Original data:\\
    The bill does not become law, unless houses of Congress vote to override the veto.\\
    \medskip
    
    Input data:\\
    \setlength\tabcolsep{1.7pt}
    \begin{tabular}{|c|c|c|c|c|c|c|c|c|c|}
    \hline
    the & bill & does & not & become & \textbf{law} & \textbf{unless} & \textbf{houses} & \textbf{of} & \textbf{congress} \\
    \hline
    \end{tabular}
    
    \begin{tabular}{|c|c|c|c|c|c|c|c|c|c|}
    \hline
    \textbf{law} & \textbf{unless} & \textbf{houses} & \textbf{of} & \textbf{congress} & vote & to & override & the & veto. \\
    \hline
    \end{tabular}
    \medskip
    
    Plain text output:\\
    \begin{tabular}{|c|c|c|c|c|c|c|c|c|c|}
    \hline
    The & bill & does & not & become & \textbf{law,} & \textbf{unless} & \textbf{houses} & \textbf{of} & \textbf{Congress} \\
    \hline
    \end{tabular}
    
    \begin{tabular}{|c|c|c|c|c|c|c|c|c|c|}
    \hline
    \textbf{law,} & \textbf{unless} & \textbf{houses} & \textbf{of} & \textbf{Congress} & vote & to & override & the & veto. \\
    \hline
    \end{tabular}
    \medskip
    
    \setlength\tabcolsep{6pt}
    Encoded output:\\
    \begin{tabular}{|c|c|c|c|c|c|c|c|c|c|}
    \hline
    U\$ & L\$ & L\$ & L\$ & L\$ & L, & L\$ & L\$ & L\$ & U\$ \\
    \hline
    L, & L\$ & L\$ & L\$ & U\$ & L\$ & L\$ & L\$ & L\$ & L. \\
    \hline
    \end{tabular}
    \caption{Data samples with chunk size of 10}
    \label{tab:data-sample}
\end{figure}

We prepared 2 formats of training data: plain text and encoded text \cite{cho2017nmt}. Both formats takes the lowercase text without punctuation as input. The plain text model, as the name suggest, provides output as plain text with punctuation and capitalization. The encoded text model, on the other hand, gives the result in an encoded format that contains only 6 classes
as showed in Table~\ref{tab:dataset}.
It is obvious that the encoded format will help the model to train and infer
faster than the plain text since its vocabulary size is fixed and very limited.
However, due to the limited vocabulary size, the decoder of the end-to-end model does not
have much information of the words and the context information.
We are interested to see how this method affect the quality in comparision with
the plain text model.

\section{Experiments and Results}
\subsection{Corpus Description}
To train and evaluate the proposed method, we use the British National Corpus (BNC) \cite{bnc} 
that contains 100 million words in both written and spoken language from a wide range of sources. It is designed to represent a large cross-section of British English from late 20\textsuperscript{th} century. We use the XML edition which contains 4049 files with the size of 515 MB in total. 
The library NLTK \cite{loper2002nltk} is used to extract 6M sentences from BNC dataset. For the test set, we use  67 thousand sentences. The number of label instances for each of the punctuation and capitalization classes available in our
training and testing data set are displayed in Table~\ref{tab:dataset}.

\begin{table}[th]
  \caption{BNC dataset detail. ``U'' and ``L'' respectively denote uppercase and lowercase word (either first or all character); ``.'', ``,'' and ``?'' denotes full stop, comma, and question mark.
The dollar sign (``\$'') indicates there are no punctuation coming after the word.}
  \label{tab:dataset}
  \centering
  \begin{tabular}{ r  r  r}
    \toprule
    \multicolumn{1}{c}{\textbf{Class}} &
    \multicolumn{1}{c}{\textbf{Training}} &
    \multicolumn{1}{c}{\textbf{Testing}} \\
    \midrule
    U  & 13 M   & 146 K    \\
    L  & 81 M   & 1 M   \\
    .  & 4.6 M    & 54 K    \\
    ,  & 4.9 M    & 57 K    \\
    ?  & 380 K      & 5 K    \\
    \$ & 87 M   & 1 M   \\
    \bottomrule
  \end{tabular}
\end{table}

\subsection{Evaluation metric}
The models (described in section 3.1) are evaluated using precision, recall, and $F_1$ scores. 
For ease of representation, we converted output words and punctuation to the 6-class encoded format as illustrated in Table~\ref{tab:dataset}.
The evaluation results indicate how well the method can predict
truecase of words and punctuation restoration. Since prediction of 
lowercase and blank space
are good in every models, we ignore them in compare table. 

\subsection{Evaluation of chunk-merging}

\begin{figure}[t]
  \centering
  \includegraphics[width=\linewidth]{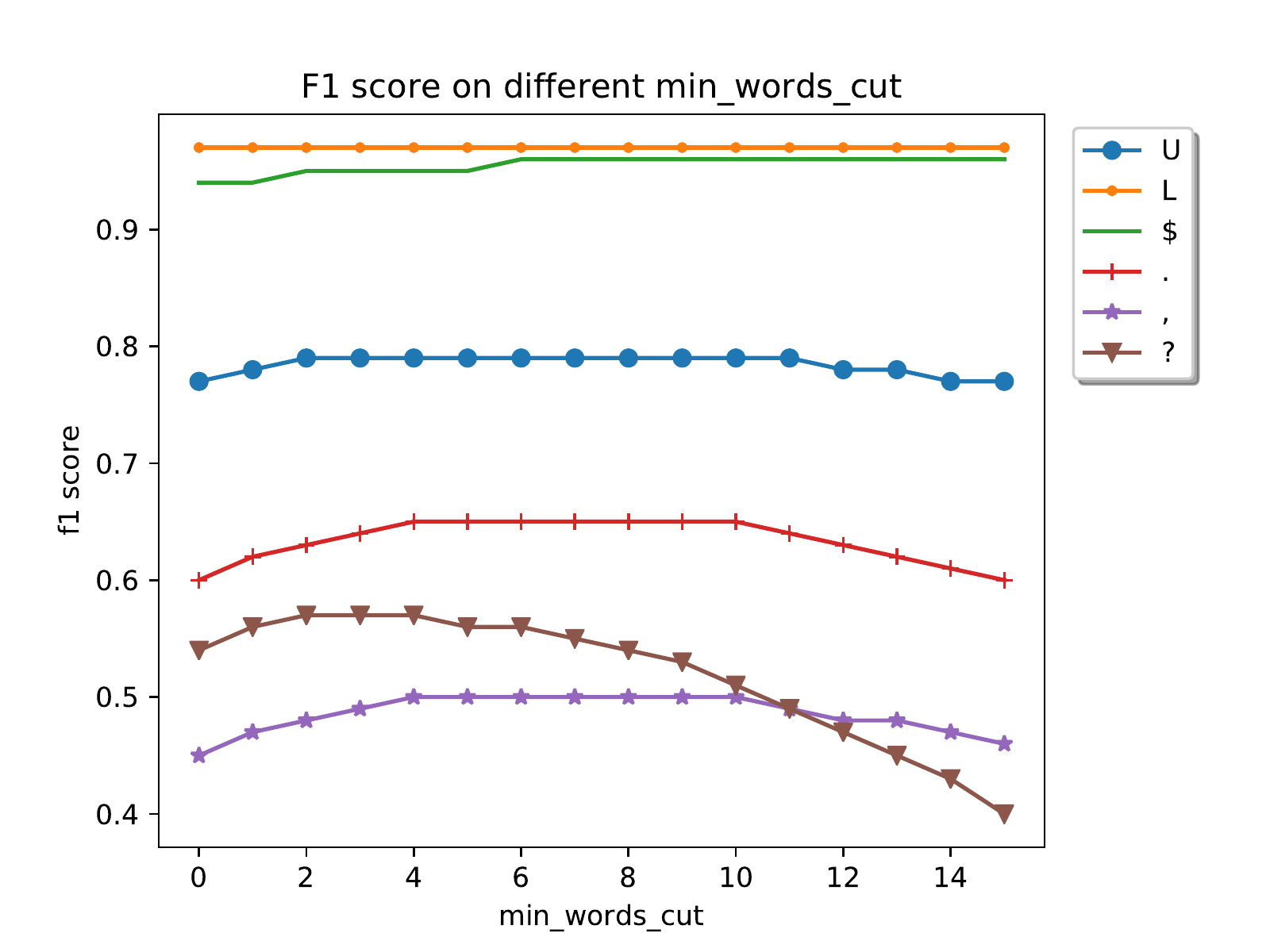}
  \caption{F1-score on different min\_word\_cut. It  peak  in  the  middle range of overlap size (4-10). Predicting uppercase  and  lowercase  are  stable  and  independent  from min\_word\_cut, question mark is quite sensitive with this hyper-parameter.}
  \label{fig:f1-min-words-cut}
\end{figure}


\begin{table}[th]
  \setlength\tabcolsep{4pt}
  \caption{Comparison Seq2seq LSTM with and without using Chunk Merging for plain text format}
  \label{tab:results-overlap-lstm}
  \centering
  \begin{tabular}{r | r r r r}
    \toprule
    \multicolumn{1}{c}{\textbf{Model}} &
    \multicolumn{1}{c}{\textbf{Class}} &
    \multicolumn{1}{c}{\textbf{Precision}} &
    \multicolumn{1}{c}{\textbf{Recall}} &
    \multicolumn{1}{c}{\textbf{F1-score}} \\
    \midrule




\multirow{4}{*}{\shortstack[r]{Chunk Merging \\ Seq2seq LSTM}} 
    &  U  &  0.74 &     0.53 &     \textbf{0.62} \\
    &  .  &  0.43    &  0.41   &   \textbf{0.42} \\
    &  ,  &  0.10    &  0.87   &   \textbf{0.19}  \\
    &  ?  &  0.49   &   0.22   &   \textbf{0.30} \\
    \hline



\multirow{4}{*}{\shortstack[r]{Non-Chunk Merging \\ Seq2seq LSTM}} 
    &  U  &  0.70 &     0.53 &     0.61 \\
    &  .  &  0.40    &  0.41   &   0.41 \\
    &  ,  &  0.10    &  0.85   &   0.18  \\
    &  ?  &  0.45   &   0.20   &   0.28 \\
    \bottomrule
  \end{tabular}
\end{table}

\begin{table}[th]
  \setlength\tabcolsep{4pt}
  \caption{Comparison Evolved Transformer with and without using Chunk Merging for plain text format}
  \label{tab:results-overlap-transformer}
  \centering
  \begin{tabular}{r | r r r r}
    \toprule
    \multicolumn{1}{c}{\textbf{Model}} &
    \multicolumn{1}{c}{\textbf{Class}} &
    \multicolumn{1}{c}{\textbf{Precision}} &
    \multicolumn{1}{c}{\textbf{Recall}} &
    \multicolumn{1}{c}{\textbf{F1-score}} \\
    \midrule

\multirow{4}{*}{\shortstack[r]{Chunk Merging \\ Evolved Transformer}} 
    &  U  &  0.90 &     0.84 &     \textbf{0.87} \\
    &  .  &  0.74    &  0.72   &   \textbf{0.73} \\
    &  ,  &  0.61    &  0.51   &   \textbf{0.56}  \\
    &  ?  &  0.82   &   0.63   &  \textbf{0.71} \\
    \hline


\multirow{4}{*}{\shortstack[r]{Non-Chunk Merging \\ Evolved Transformer}} 
    &  U  &  0.84 &     0.79 &     0.81 \\
    &  .  &  0.56    &  0.66   &   0.61 \\
    &  ,  &  0.40    &  0.42   &   0.41  \\
    &  ?  &  0.70   &   0.46   &  0.56 \\




    \bottomrule
  \end{tabular}
\end{table}

Table~\ref{tab:results-overlap-lstm} shows the result of
the seq2seq LSTM model with and without chunk-merging.
As we can see, with the help of chunk merging, $F_1$
score on all classes are improved consistently by 1\%.
The result indicates that the overlapped words
give the model more information to make better
prediction, and that our
chunk-merging method can select good portion of
the overlap area.

\begin{figure}[t]
  \centering
  \includegraphics[width=\linewidth]{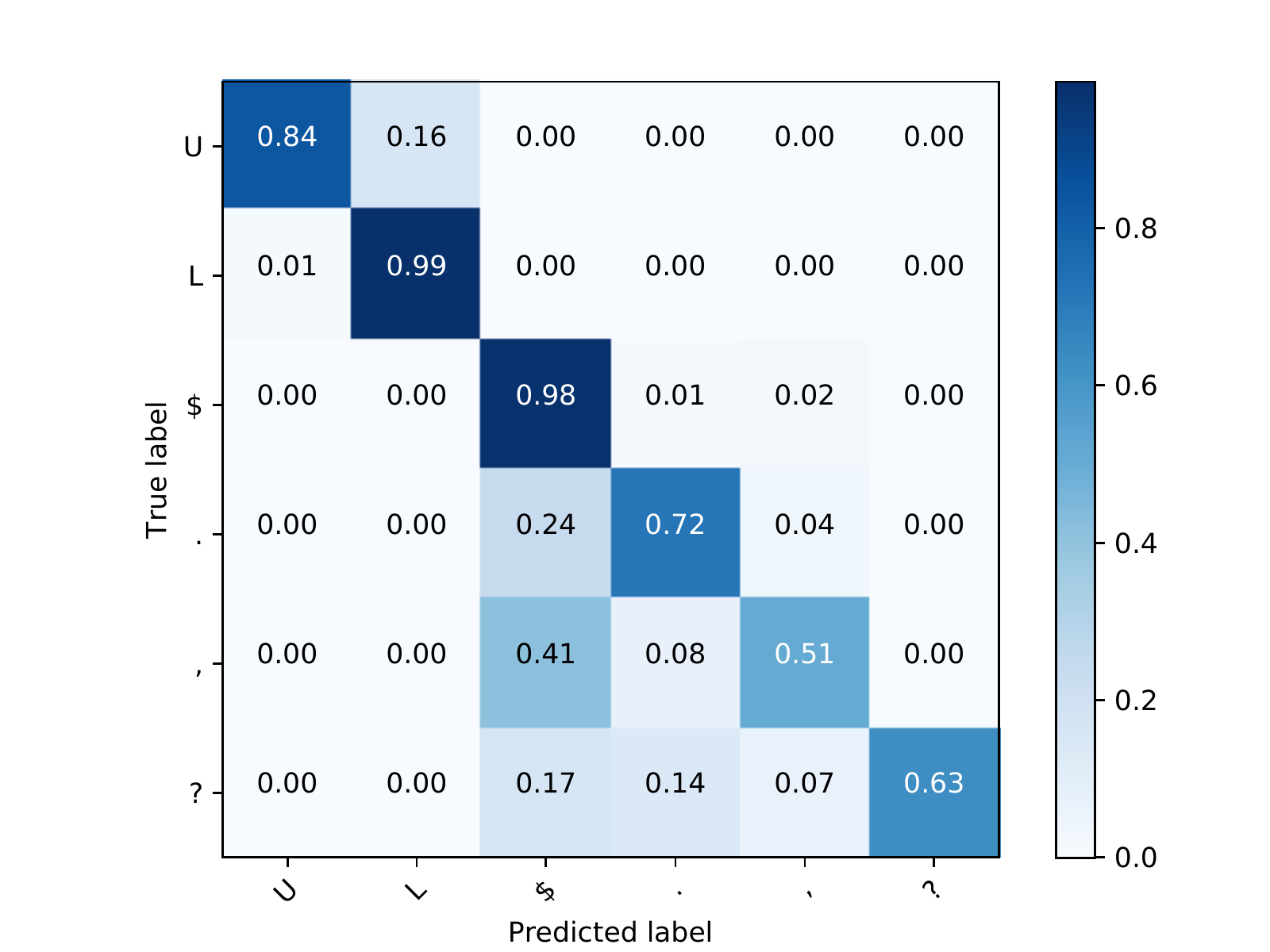}
  \caption{Confusion matrix of Evolved Transformer model with plain text and overlapping format}
  \label{fig:confusion}
\end{figure}

The chunk-merging method even shows superior performance over non-chunk-merging when it is used with
Evolved Transformer models. Results on Table~\ref{tab:results-overlap-transformer} shows that the
prediction accuracy of the question mark raises from
56\% to 71\%, this is a margin of 15\% improvement and the
minimum improvement of the system is 6\% for the uppercase class.
Figure~\ref{fig:confusion} displays the confusion matrix of the model. The matrix shows that
the comma is the most difficult class to predict and it is often mis-predicted
as blank characters. In addition, the matrix also indicates that the model
always predict a word (either lowercase or uppercase) when the input is word.

The results prove our hypothesis that there is not enough context for model to predict efficiently at the beginning and the end of each sample and that drawback can be overcome by adding more context
with chunk overlapping and chunk-merging method.

\subsection{Evaluation on plain-text model and encoded-text model}
We further compare the result on models using plain text and encoded text. The ones with plain text outperform the ones with encoded text, however the model using encoded text has smaller model size and is faster for inference. The details are in Table~\ref{tab:results-encoded-plain}

\begin{table}[th]
  \setlength\tabcolsep{4pt}
  \caption{Comparison of results encoded text and plain text using Evolved Transformer}
  \label{tab:results-encoded-plain}
  \centering
  \begin{tabular}{r | r r r r}
    \toprule
    \multicolumn{1}{c}{\textbf{Model}} &
    \multicolumn{1}{c}{\textbf{Class}} &
    \multicolumn{1}{c}{\textbf{Precision}} &
    \multicolumn{1}{c}{\textbf{Recall}} &
    \multicolumn{1}{c}{\textbf{F1-score}} \\
    \midrule
\multirow{4}{*}{\shortstack[r]{Encoded Text \\ Chunk Merging \\ Evolved Transformer}}
    &  U  &  0.87 &     0.80 &     0.84 \\
    &  .  &  0.68    &  0.66   &   0.67 \\
    &  ,  &  0.50    &  0.40   &   0.44  \\
    &  ?  &  0.76   &   0.55   &   0.63 \\
    \hline

\multirow{4}{*}{\shortstack[r]{Plain Text \\ Chunk Merging \\ Evolved Transformer}} 
    &  U  &  0.90 &     0.84 &     \textbf{0.87} \\
    &  .  &  0.74    &  0.72   &   \textbf{0.73} \\
    &  ,  &  0.61    &  0.51   &   \textbf{0.56}  \\
    &  ?  &  0.82   &   0.63   &   \textbf{0.71} \\
    \bottomrule
  \end{tabular}
\end{table}

To explore the impact of \verb|min_words_cut| value to the quality of the result, we performed the experiment on sequence-to-sequence LSTM  model with the overlapping of 15 words and \verb|min_words_cut| ranges from 0 to 15. The outcome shown in Figure~\ref{fig:f1-min-words-cut} indicates that f1-scores peak in the middle range of chunk size (4-10). It demonstrate that predictions of uppercase and lowercase are stable and independent from \verb|min_words_cut|.

As processing chunks is paralleled and the concatenation algorithm has $\mathcal{O}(n)$, this approach is fast and proved to be superior to conventional methods.

\section{Conclusion}

In this research, we have proposed an end-to-end model that restores both punctuation and capitalization in one go. With chunk-split-merging,
the method can splits and processes sentences in parallel and merges outputs to form the final sentence output. Experiments shows that the approach outperform existing methods that do not utilize chunk-merging by a significant margin, especially when combining with Evolved Transformer. In the future, we will integrate this solution with ASR model to form an end-to-end model that can transform speech to a well format text document.



\bibliography{mybib}

\begin{thebibliography}{10}
\providecommand{\url}[1]{#1}
\csname url@samestyle\endcsname
\providecommand{\newblock}{\relax}
\providecommand{\bibinfo}[2]{#2}
\providecommand{\BIBentrySTDinterwordspacing}{\spaceskip=0pt\relax}
\providecommand{\BIBentryALTinterwordstretchfactor}{4}
\providecommand{\BIBentryALTinterwordspacing}{\spaceskip=\fontdimen2\font plus
\BIBentryALTinterwordstretchfactor\fontdimen3\font minus
  \fontdimen4\font\relax}
\providecommand{\BIBforeignlanguage}[2]{{%
\expandafter\ifx\csname l@#1\endcsname\relax
\typeout{** WARNING: IEEEtran.bst: No hyphenation pattern has been}%
\typeout{** loaded for the language `#1'. Using the pattern for}%
\typeout{** the default language instead.}%
\else
\language=\csname l@#1\endcsname
\fi
#2}}
\providecommand{\BIBdecl}{\relax}
\BIBdecl

\bibitem{cho2012segmentation}
E.~Cho, J.~Niehues, and A.~Waibel, ``Segmentation and punctuation prediction in
  speech language translation using a monolingual translation system,'' in
  \emph{International Workshop on Spoken Language Translation (IWSLT) 2012},
  2012.

\bibitem{tkachenko2012named}
M.~Tkachenko and A.~Simanovsky, ``Named entity recognition: Exploring
  features.'' in \emph{Proceeding of KONVENS}, 2012, pp. 118--127.

\bibitem{yadav2018survey}
V.~Yadav and S.~Bethard, ``A survey on recent advances in named entity
  recognition from deep learning models,'' in \emph{Proceedings of CICLing},
  2018, pp. 2145--2158.

\bibitem{_elasko_2018}
\BIBentryALTinterwordspacing
P.~Żelasko, P.~Szymański, J.~Mizgajski, A.~Szymczak, Y.~Carmiel, and
  N.~Dehak, ``Punctuation prediction model for conversational speech,''
  \emph{Interspeech 2018}, Sep 2018. [Online]. Available:
  \url{http://dx.doi.org/10.21437/Interspeech.2018-1096}
\BIBentrySTDinterwordspacing

\bibitem{tilk2015lstm}
O.~Tilk and T.~Alum{\"a}e, ``Lstm for punctuation restoration in speech
  transcripts,'' in \emph{Sixteenth annual conference of the international
  speech communication association}, 2015.

\bibitem{lu2010better}
W.~Lu and H.~T. Ng, ``Better punctuation prediction with dynamic conditional
  random fields,'' in \emph{Proceedings of the 2010 conference on empirical
  methods in natural language processing}, 2010, pp. 177--186.

\bibitem{ueffing2013improved}
N.~Ueffing, M.~Bisani, and P.~Vozila, ``Improved models for automatic
  punctuation prediction for spoken and written text.'' in \emph{Interspeech},
  2013, pp. 3097--3101.

\bibitem{lafferty2001conditional}
J.~Lafferty, A.~McCallum, and F.~C. Pereira, ``Conditional random fields:
  Probabilistic models for segmenting and labeling sequence data,'' 2001.

\bibitem{cho2017nmt}
E.~Cho, J.~Niehues, and A.~Waibel, ``Nmt-based segmentation and punctuation
  insertion for real-time spoken language translation.'' in \emph{INTERSPEECH},
  2017, pp. 2645--2649.

\bibitem{opennmt}
\BIBentryALTinterwordspacing
G.~Klein, Y.~Kim, Y.~Deng, J.~Senellart, and A.~M. Rush, ``Open{NMT}:
  Open-source toolkit for neural machine translation,'' in \emph{Proc. ACL},
  2017. [Online]. Available: \url{https://doi.org/10.18653/v1/P17-4012}
\BIBentrySTDinterwordspacing

\bibitem{bahdanau2014neural}
D.~Bahdanau, K.~Cho, and Y.~Bengio, ``Neural machine translation by jointly
  learning to align and translate,'' 2014.

\bibitem{sutskever2014sequence}
I.~Sutskever, O.~Vinyals, and Q.~V. Le, ``Sequence to sequence learning with
  neural networks,'' in \emph{Advances in neural information processing
  systems}, 2014, pp. 3104--3112.

\bibitem{vaswani2017attention}
A.~Vaswani, N.~Shazeer, N.~Parmar, J.~Uszkoreit, L.~Jones, A.~N. Gomez,
  {\L}.~Kaiser, and I.~Polosukhin, ``Attention is all you need,'' in
  \emph{Advances in Neural Information Processing Systems}, 2017, pp.
  5998--6008.

\bibitem{zoph2016neural}
B.~Zoph and Q.~V. Le, ``Neural architecture search with reinforcement
  learning,'' \emph{arXiv preprint arXiv:1611.01578}, 2016.

\bibitem{tensor2tensor}
\BIBentryALTinterwordspacing
A.~Vaswani, S.~Bengio, E.~Brevdo, F.~Chollet, A.~N. Gomez, S.~Gouws, L.~Jones,
  L.~Kaiser, N.~Kalchbrenner, N.~Parmar, R.~Sepassi, N.~Shazeer, and
  J.~Uszkoreit, ``Tensor2tensor for neural machine translation,'' \emph{CoRR},
  vol. abs/1803.07416, 2018. [Online]. Available:
  \url{http://arxiv.org/abs/1803.07416}
\BIBentrySTDinterwordspacing

\bibitem{bnc}
B.~Consortium, \emph{The British National Corpus, version 3 (BNC XML
  Edition)}.\hskip 1em plus 0.5em minus 0.4em\relax Bodleian Libraries,
  University of Oxford, 2007.

\bibitem{loper2002nltk}
E.~Loper and S.~Bird, ``Nltk: the natural language toolkit,'' \emph{arXiv
  preprint cs/0205028}, 2002.

\end{thebibliography}

\end{document}